\documentclass[journal, 9pt]{IEEEtran}
\usepackage[pdftex]{graphicx}

\usepackage[cmex10]{amsmath}
\usepackage{array}
\usepackage{fixltx2e}
\usepackage{enumerate}
\usepackage{multicol}
\usepackage{multirow}
\usepackage{url}
\usepackage{subcaption}
\usepackage{algorithm} 
\usepackage{algorithmic}
\usepackage{arydshln}
\usepackage{multirow}
\usepackage{verbatim}
\usepackage{rotating}
\usepackage{textcomp}

\usepackage{mathptmx} 
\usepackage{threeparttable}
\usepackage{cite}
\usepackage{array}
\usepackage[usenames,dvipsnames]{color, colortbl}
\definecolor{Gray}{gray}{0.9}
\definecolor{LightCyan}{rgb}{0.88,1,1}
\definecolor{ulascolor}{rgb}{0.6, 0.3, 0.2}
\definecolor{dark_purple}{rgb}{0.4, 0.0, 0.4}
\usepackage{amssymb}
\usepackage{amsfonts}
\usepackage{amsmath}

\begin{document}
\title{A Novel Extension to Fuzzy Connectivity for Body Composition Analysis: Applications in Thigh, Brain, and Whole Body Tissue Segmentation}
\author{Ismail~Irmakci,
Sarfaraz~Hussein,
        Aydogan Savran,
		Rita~R. Kalyani,
		David~Reiter,
        Chee W. Chia,
        Kenneth W. Fishbein,
        Richard G. Spencer,
        Luigi Ferrucci,
        and~Ulas~Bagci$^*$,~\IEEEmembership{Senior Member,~IEEE,}
\thanks{\emph{Asterisk indicates corresponding author.} e-mail: ulasbagci@gmail.com}
\thanks{I. Irmakci and A. Savran are with Ege University, Izmir, Turkey. Irmakci's research is supported by scientific council of Turkey (TUBITAK-BIDEB 2214/A). }
\thanks{C.W. Chia, K.W. Fishbein, R.G. Spencer, L. Ferrucci and D. Reiter are with the National Institute on Aging, National Institutes of Health (NIH), Baltimore, MD.}
\thanks{R. Kalyani is with the Department of Medicine, Johns Hopkins University School of Medicine, Baltimore, MD.}
\thanks{U. Bagci and S. Hussein are with Center for Research in Computer Vision at University of Central Florida, Orlando, FL.}
}

\markboth{IEEE Transactions on Biomedical Engineering,~Vol.~xxx, No.~xxx, Month~2018}%
{Irmakci \MakeLowercase{\textit{et al.}}: Multicontrast Thigh MRI Segmentation}

\maketitle

\begin{abstract}
Magnetic resonance imaging (MRI) is the non-invasive modality of choice for body tissue composition analysis due to its excellent soft tissue contrast and lack of ionizing radiation. However, quantification of body composition requires an accurate segmentation of fat, muscle and other tissues from MR images, which remains a challenging goal due to the intensity overlap between them. In this study, we propose a fully automated, data-driven image segmentation platform that addresses multiple difficulties in segmenting MR images such as varying inhomogeneity, non-standardness, and noise, while producing high-quality definition of different tissues. In contrast to most approaches in the literature, we perform segmentation operation by combining three different MRI contrasts and a novel segmentation tool which takes into account variability in the data. The proposed system, based on a novel affinity definition within the fuzzy connectivity (FC) image segmentation family, prevents the need for user intervention and reparametrization of the segmentation algorithms. In order to make the whole system fully automated, we adapt an affinity propagation clustering algorithm to roughly identify tissue regions and image background. We perform a thorough evaluation of the proposed algorithm's individual steps as well as comparison with several approaches from the literature for the main application of muscle/fat separation. Furthermore, whole-body tissue composition and brain tissue delineation were conducted to show the generalization ability of the proposed system. This new automated platform outperforms other state-of-the-art segmentation approaches both in accuracy and efficiency.
\end{abstract}


\section{Introduction}
\begin{figure}
\includegraphics[width=9 cm]{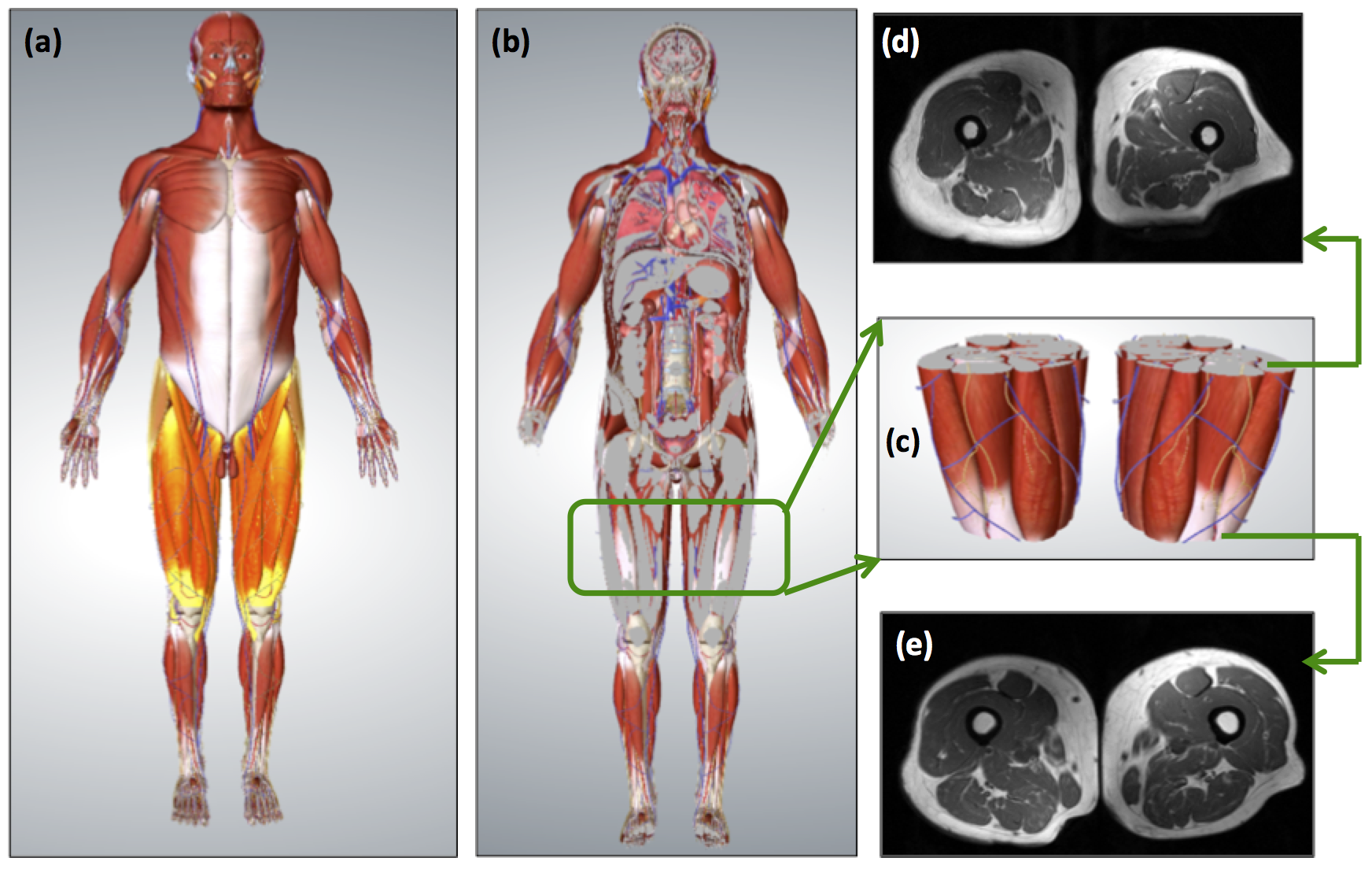}
\caption{Fat and muscle are important outcome measures in the evaluation of several health conditions. (a) depicts the muscular representation of the whole-body where thigh regions are highlighted. (b) is the coronal view of the whole-body representation which shows bone, fat, and muscle distribution in thigh region (zoomed version is in (c)). Pre-defined thigh region (c), (d) and (e) show MRI correspondences of the starting and ending positions of the thigh. MRI is a non-invasive imaging modality, allowing visualization of tissues with its excellent soft tissue contrast and optimal discrimination between different tissue types.\label{fig:motivation}}
\end{figure}

\IEEEPARstart{B}{ody} composition changes are observed in aging and may be related to the development of disorders such as metabolic syndrome and diabetes. Sarcopenia, the progressive loss of muscle mass with age tends to be associated with higher amount of fat (adipose) in the thigh muscle (Figure~\ref{fig:motivation}). It is noteworthy that the rate of decline in muscle strength with age exceeds what would be predicted based on muscle mass alone. This suggests that perhaps age affects muscle quality by facilitating fat infiltration. These facts propel current investigations, and research into the non-invasive quantification of body tissue composition is the subject of a great deal of current research~\cite{sarfaraz}.
Magnetic resonance imaging (MRI) provides excellent soft tissue contrast (Figure~\ref{fig:motivation}) without exposing patients to ionizing radiation, making it the preferred modality for body composition studies\footnote{Figure was created with the BioDigital Human Visualization Platform \textcopyright.}. Previously, we developed segmentation methods for abdominal, thoracic, and brown fat quantification~\cite{sarfaraz}. Here, we focus on fat/muscle quantification in thigh regions using multiple MRI image contrasts.

\subsection{Unmet clinical need and summary of challenges} 
Conventional methods of manual or semi-automated segmentation methods either fail to provide high sensitivity and specificity or require substantial user interaction with the segmentation software. Because these methods are time consuming and have limited inter-raters reliability, they cannot be used for large-scale studies including those for routine clinical evaluations. There are other difficulties such as noise, inhomogeneity, and variable intensity that are not completely avoidable in MRI methodology that may affect any image-based quantification task (i.e., this is specifically called ``intensity non-standardness"). It is highly desirable to have an automated method that can create standardized thigh MR images where non-anatomical uncertainties are minimized while providing accurate and efficient quantification of fat and muscle tissue compartments in the thigh.

To the best of our knowledge, there is no algorithm reported to conduct fat and muscle segmentation within a very short time interval (i.e., seconds) and with high precision. Efficiency is of particular importance because it may be one of the major reasons that many algorithms are not adopted into clinical settings or large research studies. Last, but not least, benefits of multi-contrast MRI have not been fully explored for fat and muscle quantification as yet. Here, we prove the complementary nature of multi-contrast imaging quantitatively. 

\subsection{Background and related works}
Prescott et al.~\cite{prescott2011anatomically} used level set segmentation methods by utilizing various initialization templates to perform fat/muscle segmentation of thigh MR images. This idea was built upon the success of atlas-based segmentation methods in brain images and anatomical organs where tissue distribution played a significant role~\cite{okada2008automated,aljabar2009multi}. 
In another study, Ahmad et al.~\cite{ahmad2014atlas} performed a semi-automated segmentation of quadriceps muscles in thigh MRI. 
In addition to requiring user input, this method relied heavily on proper segmentation from a single slice and carefully defined prior information. 

Shape models were also extensively reported in the literature, particularly when tissue boundary separation was difficult. Essafi et al.'s~\cite{essafi2009wavelet} study can be categorized under this setting. The proposed algorithm used diffusion wavelets and geometric constraints to optimize the position of anatomical landmarks which helped separation of different tissues. The method achieved superior performance in modeling the heterogeneity of the muscle topology. However, it is unknown how this method will perform under different MRI inhomogeneities and noise, which were not tested in their study. In a similar fashion, the same objective of modeling tissue heterogeneity was addressed in \cite{wang20103d}, where a point distribution model along with higher order Markov random field (MRF) model were used to represent tissue boundaries. 
Yet, the performances of such methods often suffered when boundary points were not localized reliably. Moreover, segmentation outcome was extremely sensitive to initialization of the model assembly, as reported also in our previous works~\cite{chen2012automatic,bagci2012hierarchical}.

Recent investigations in this field have sought to address the challenges described above using different methods. For instance, Andrews et al.~\cite{andrews2015generalized}, included a generalized log-ratio notion in their quantification algorithm to represent prior information of anatomical volume and tissue adjacency in a probabilistic space. Random decision forest algorithm was used to learn intensity and texture characteristics of muscle and fat tissues. Later, this information was combined with a shape model in order to perform the final segmentation. 
Makrogiannis et al.~\cite{makrogiannis2016image} used a Gaussian mixture model (GMM) for modeling the muscle, fat, and inter-muscle adipose tissue (IMAT) jointly from water and fat-suppressed MRI images. Due to the nature of this problem, clustering-based approaches were better suited for quantifying IMAT. However, clustering-based approaches were not suitable for subcutaneous fat (i.e., fat under the skin) and muscle separation, and a separate step for their delineations was required. To address this challenge, the authors used active contour-based approach with user-interventions to quantify fat and muscle distributions from multiple MR image contrasts. A subsequent study by the same authors on the same data set included a modified active contour based delineation method, which wasn't completely successful in segmenting fat/muscle in thigh regions for individual MRI modalities~\cite{makrogiannis2012automated}. 

\subsection{Overview of the proposed approach and our contributions}
\begin{figure*}
\includegraphics[width=12.5 cm]{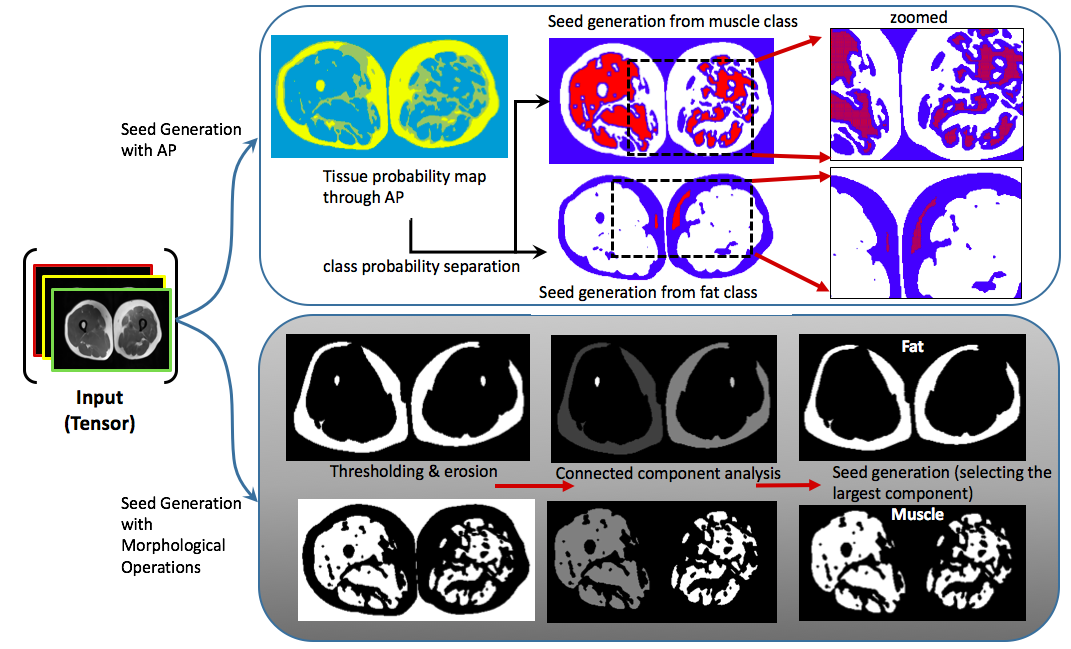}
\includegraphics[height=7.5 cm]{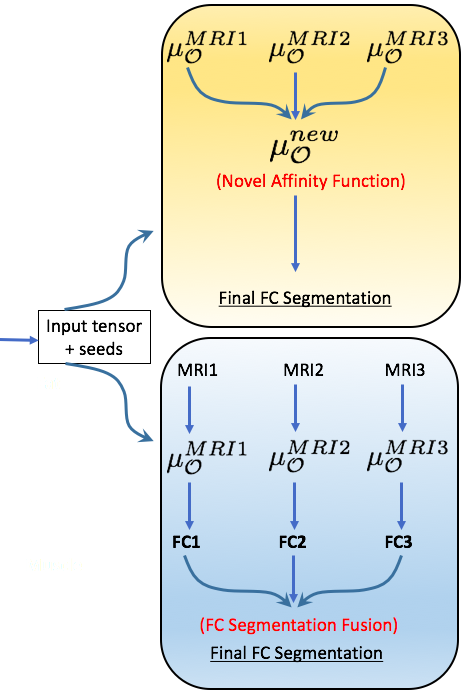}
\caption{An overview of the proposed framework is illustrated. First, the MRI images are pre-processed and cleaned using the approaches explained in Section II-B. The cleaned images are then combined to create an input tensor, which is then fed into a seed selection scheme. We have discussed two seeding strategies i.e. affinity propagation (AP) and morphology based operators. The final stage devises two alternative fusion strategies for multi-object segmentation from multi-contrast MRI: fusion at the affinity space and fusion at the decision level (last column). While affinity combination uses evidence based inference from probability density function and segmentation results of each contrast image, decision fusion uses Karnaugh map based multi-label fusion strategy to refine the final tissue segmentation.  \label{fig:overview}}
\end{figure*}

We propose a generic image segmentation method based on a novel fuzzy connectivity segmentation algorithm. Specifically, we design a novel computational platform for joint segmentation of fat and muscle from multi-contrast MR images of the thigh. Later, we show that the proposed algorithm is capable of classifying tissues in whole body MRI and brain MRIs as well. The proposed algorithm has two modules: automated seed selection (i.e., detection/localization) and delineation (i.e., precise boundary identification). Figure~\ref{fig:overview} illustrates the overview of the proposed method with its two modules for the main application in this paper: muscle/fat separation from multi-contrast MRIs. Prior to segmentation, each MR image is pre-processed, and the three image contrasts are combined to form an input tensor (i.e., a vector including inputs from multiple sources is called as MRI or input tensor from now on). 

Briefly, pre-processing includes consecutive filtering operations that remove bias-field (inhomogeneity)~\cite{interplay}, minimize noise~\cite{saha2001scale}, and stabilize intensity non-standardness (i.e., acquisition-to-acquisition signal intensity variations)~\cite{interplay,bagci2011intensity,nyul1999standardizing}. Resulting MRI images, referred to as ``clean images", are used for tissue segmentation and quantification. In the first module, we propose to use affinity
propagation (AP) (and alternatively morphology based) seed generation approaches to identify sample locations for fat and muscle tissues as well as the background. 

The second module describes our efforts for segmenting fat and muscle tissues based on sampled seeds in the first module to make the segmentation algorithm fully automated. We utilize the clean volumetric MRI in three different image contrasts jointly, and apply a novel affinity function to drive multi-object fuzzy connectivity (FC) image segmentation. This new affinity function adaptively combines individual affinities pertaining to each MRI contrast. 

\subsection*{Our contributions}
\begin{itemize}
\item Our proposed method is completely data-driven and fully automated. 
\item Unlike the most previous works, we consider multiple MRI contrasts in a unified segmentation platform to optimize complementary information from those individual contrasts. Compared to other studies that use multiple MRI contrast, our proposed study is different and unique since we perform the segmentation by combining strengths of other contrast images through a multi-object affinity function within the FC image segmentation algorithm. 
\item Seed-based segmentation algorithms use different strategies spanning from manual to automated approaches. In our implementation, we optimize the seed selection procedure without the need for extensive search. Instead, we adapt an unsupervised clustering method (i.e., AP) to roughly and efficiently identify objects of interest prior to the delineation procedure.
\end{itemize}
The remainder of this manuscript is organized as follows. Section II describes the details of the proposed method along with potential alternatives. Section III introduces experiments and performance evaluations. The manuscript ends with a discussion and concluding remarks in Section IV.

\section{Materials and Methods}
\subsection{Data and reference standards}
\textbf{Muscle/Fat separation application:} MRI were acquired using a 3T Philips Achieva MRI scanner (Philips Healthcare, Best, The Netherlands) equipped with a Q-body radiofrequency coil for transmission and reception. Three separate image volumes were obtained using a spoiled gradient echo for readout, with coverage from the proximal to distal ends of the femur using 80 slices in the foot to head direction, a field of view (FOV) of 440 x 296 x 400 mm$^3$ and a voxel size of 1 x 1 mm$^2$ in-plane, and slice thickness varies from 1 mm to 3 mm in different scans (one particular scan was with 5 mm slice thickness). Acquisition parameters included: repetition time (TR) = 7.7 ms, echo time (TE) = 2.4ms, number of signal averages (NSA) = 2, flip angle (FA) = 25 degrees, and the bandwidth of 452 Hz. The three image acquisitions were tailored to yield images containing signal from water-and-fat, fat-only, and water-only, respectively. The first image containing water-and-fat was obtained with the FFE sequence described above. For fat-only and water-only images, water and fat suppression was obtained using spectral pre-saturation with inversion recovery (SPIR). Fifty subjects' 150 MR images (three contrast images for each subject) were retrospectively collected from BLSA (Baltimore Longitudinal Study of Aging) database~\cite{BLSA}. Experiments were conducted on a regular PC with Intel Core i5 at 2.9 GHz configuration with 16 GB RAM at 1867 MHz speed. The details about the experimental cohort including age, weight, height, and BMI of the subjects are included in Table~\ref{table:data}.

\textbf{Whole-Body tissue quantification:} Whole-body MR imaging was conducted using 3-D dual-echo, spoiled gradient echo sequences with Dixon-type fat-water separation (Wollenweber et al.~\cite{wollenweber2013evaluation}). The following protocol was used for determining the images. FOV was 48 cm x 48 cm, TE1/TE2 was 1.15 ms / 2.3 ms, FA/TR=$12^{o}/4.33$ ms, acquisition matrix = 256 x 128 with 3.8 mm slices, and volumetric images of in-phase, out-phase, fat, and water were reconstructed using Dixon/IDEAL processing on the GE MR systems with a breath hold acquisition with a target of under 20 s. We have used 3 whole body MRI scans to do our segmentation feasibility study.

\textbf{Brain tissue delineation:} For brain tissue segmentation, the publicly available BRAINWEB dataset was used in our experiments. The dataset comprises simulated phantom images which were constructed from a high-resolution (1-mm isotropic voxels) data which was generated after registering T1-weighted scans having gradient-echo acquisitions with TR/TE/FA $=$ 18 ms/10 ms/30$^{\circ}$. The specificity and sensitivity results of the whole body and brain tissue segmentations are given in Table IV in results section.

\begin{table}
\caption{Details about the experimental cohort used in the experiments with age (years), weight (kg), height (cm) and Body Mass Index (BMI).} \label{table:data}
{\begin{tabular}{ |p{2cm}||p{2.5cm}|p{2.5cm}|p{1cm}|  }
 \hline
 \multicolumn{3}{|c|}{\textbf{Experimental Cohort}} \\
 \hline
\textbf{Characteristics} &  \textbf{Range} & \textbf{mean $\pm$ std.}\\
 \hline
 Age   & [44, 89] years    &71 $\pm$ 11 years\\
Weight&   [47.7, 121] kg  & 77.5 $\pm$ 17  kg\\
Height&   [148, 187.1] cm  & 169.32 $\pm$ 9.26 cm\\
BMI&   [18.67, 45.68]  & 26.87 $\pm$ 5.02\\
 \hline
\end{tabular}}
\end{table}

\subsection{Pre-processing MR images}
MR images exhibit (1) acquisition-to-acquisition signal intensity variations (called intensity non-standardness), (2) inherent noise, and (3) intensity non-uniformity (i.e., bias-field or inhomogeneity). In our earlier work, we have shown that for precise quantification, all three of these signal variations should be corrected ~\cite{bagcispie,bagci2012hierarchical,bagci2011intensity,bagci2010influence,bagci2010role}. 
As the bias correction promotes noise, denoising filters should be used after the inhomogeneity correction step. Hence, the last step is the intensity standardization for correcting non-standardness among images~\cite{bagci2012hierarchical}. 

\subsubsection{Inhomogeneity correction}  Inhomogeneity occurs due to low spatial frequency intensity variations. Many MRI vendors include simple inhomogeneity correction methods during image acquisition process~\cite{axel1987intensity, bagci2010role, bagci2011intensity,bagci2010influence}. However, those methods are often based on physical phantoms that are not truly representative of patient anatomy and position. In our experiments, we use post-processing filtering method based on nonparametric nonuniform intensity normalization, called N4ITK~\cite{tustison2010n4itk}, which maximizes the high-frequency content of the tissue intensity distribution. We use the following parameters in N4ITK for reproducible results: order of spline (for fitting)=3, number of histogram bins=200, full width at half maximum for Gaussian convolution parameter=0.15.

\subsubsection{Denoising} Noise is ubiquitous in MRI scans. In addition to inherent MRI noise, image enhancement algorithms such as inhomogeneity correction can further intensify the noise. To this end, we have used the ball-scale based diffusive filtering method that preserves boundary sharpness and fine structures as reported in~\cite{bagci2012hierarchical,bagcispie,saha2001scale}, which is found to have superior performance than anisotropic diffusion based filtering methods ~\cite{saha2001scale,bagci2010ball}. 

\subsubsection{Intensity standardization} Due to acquisition-to-acquisition signal intensity variations, MR image intensities do not possess a tissue-specific numeric meaning between subsequent scans even for images acquired from the same subject, scanner, body region, or pulse sequence~\cite{nyul1999standardizing}. This phenomenon is called non-standardness. Non-standardness has been largely ignored in the literature until recently since many vendors have started to offer various intensity standardization algorithms for their image reconstruction processes~\cite{robitaille2012tissue,jager2007whole,nyul2000new}. Basically, these algorithms (called intensity standardization) are pre-processing techniques, which map image intensities into a standard intensity scale. The mapping is done through a piece-wise linear function which is obtained from a training step. In the current study, we have followed the intensity standardization approach as reported in~\cite{nyul2000new}. For reproducible results, we set the min and max intensity values into 1 and 4095, respectively, as suggested by~\cite{nyul1999standardizing}. In addition, we set only a single foreground histogram mode to be determined automatically from the imaging data.

\begin{figure*}
\centering
\includegraphics[width=13 cm]{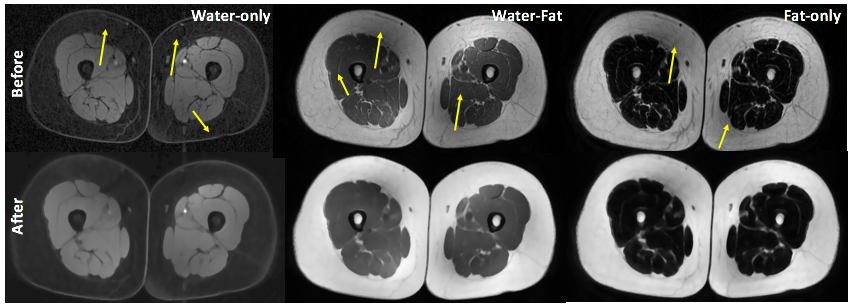}
\caption{First row includes water-only, water-fat, and fat-only MR images with inherent noise and bias-field (marked with yellow arrows). The second row shows cleaned MRI correspondences after the preprocessing steps: inhomogeneity correction, denoising, and intensity standardization. We trained standardization parameters (i.e., histogram landmarks) separately for different image modality. 
\label{fig:preprocessed}}
\end{figure*}

Figure~\ref{fig:preprocessed} shows water-only, water-fat, and fat-only MR images obtained directly from the scanner in the first row. Images include inhomogeneity and noise distortion as expected. In the second row, MR images have been cleaned through the pre-processing steps in this order: inhomogeneity correction, denoising, and intensity standardization. Substantial improvements in the images can be readily observed in clean images. Yellow arrows show inhomogeneity and noisy regions prior to pre-processing step.
As a quantitative example, we have also provided the coefficient of variance (CoV) values for each image before and after pre-processing steps in Figure~\ref{fig:cov_comparision} of the experiments and results section. As noted, the pre-processing approach improves not only visual interpretation but also reduces intensity variation for each tissue. 

\subsection{Background on FC image segmentation}
\label{subsec:fuzzy}
FC defines the ``hanging togetherness'' between any two voxels $p$ and $q$ within an image~\cite{FC}. A binary adjacency relationship ($\mu_\alpha$) determines adjacent voxels in $\alpha$-adjacency. For any two adjacent voxels $p_i$ and $p_{i+1}$, their local hanging togetherness is generally defined using an affinity function $\mu_k\left(p_i,p_{i+1}\right)$. This allows us to create probability maps when one of the voxel's label is known. When two voxels are more closely related, their affinity becomes greater and this can be used for labeling purposes (i.e., segmentation). To generalize this concept from adjacent two voxels to a set of voxels, FC considers a ``path". A path $\pi$ between $p$ and $q$ is identified as a sequence of adjacent voxels $\pi=\langle p_0=p, p_1,\dots,p_l=q \rangle$, where for an arbitrary path $\pi$, the strength of the path is defined as the ``minimum affinity" along that path:
\begin{equation}
\mu\left(\pi\right)=\mathop{\min}\limits_{0\le i\le l}\mu_k\left(p_i,p_{i+1}\right).
\end{equation}
Assuming $\mathcal{P}\left(p,q\right)$ is the set of all possible paths between $p$ and $q$, then FC between them is the strength of the strongest path (indicating closely connected voxels with strong affinities as belonging to the same object):
\begin{equation}
\mu_\kappa\left(p,q\right)=\mathop{\max}\limits_{\pi\in\mathcal{P}\left(p,q\right)}\mu\left(\pi\right).
\end{equation}

Defining a proper affinity function for two adjacent voxels is the key for a successful FC operation. Generally, the affinity function consists of three components: adjacency-based affinity $\mu_d$, homogeneity-based affinity ($\mu_\psi$), and object-based affinity ($\mu_\phi$) which are defined as:
\begin{equation}
\mu_k\left(p_i,p_{i+1}\right)=\mu_d\left(p_i,p_{i+1}\right)\sqrt{\mu_\psi\left(p_i,p_{i+1}\right)\mu _\phi\left(p_i,p_{i+1}\right)}.
\end{equation}
A wide range of mathematical functions can be used for affinities \cite{ciesielski2012fuzzy}. Herein, we use Euclidean distance for $\mu_d$, and adopt the following form of $\mu_\psi$ and $\mu_\phi$ for representing tissue class intensity properties:
\begin{equation}
\mu_\psi\left(p_i,p_{i+1}\right)=\exp\left(-\frac{\left|f\left(p_i\right)-f\left(p_{i+1}\right)\right|^2}{2\sigma _\psi^2}\right),
\end{equation}
and
\begin{multline}
\mu_\phi\left(p_i,p_{i+1}\right)= \\ \min \left(\exp\left(-\frac{\left|f\left(p_i\right)-m\right|^2}{2\sigma _\psi^2}\right), \exp\left(-\frac{\left|f\left(p_{i+1}\right)-m\right|^2}{2\sigma _\phi^2}\right)\right),
\end{multline}
where $\sigma_\psi$ and $\sigma_\phi$ control the variance, and $m$ controls the expected mean intensity of the target object. In our implementation, these values are all determined in the training step.

FC segmentation is obtained by generating a fuzzy object $\mathcal{O}$ corresponding to a set of seed points $s\in\mathcal{S}$. The fuzzy object membership value at a voxel $p$ is then determined as the maximum FC value of all seed points:
\begin{equation}
\mu_\mathcal{O}\left(p\right)=\mathop{\max}\limits_{s\in\mathcal{S}}\mu_\kappa\left(p,s\right).
\end{equation}
The final object is derived from a probability map (i.e., the fuzzy object $\mathcal{O}$) by thresholding.

\subsection{Novel Affinity Function for FC}
\begin{figure}
\includegraphics[width=9 cm]{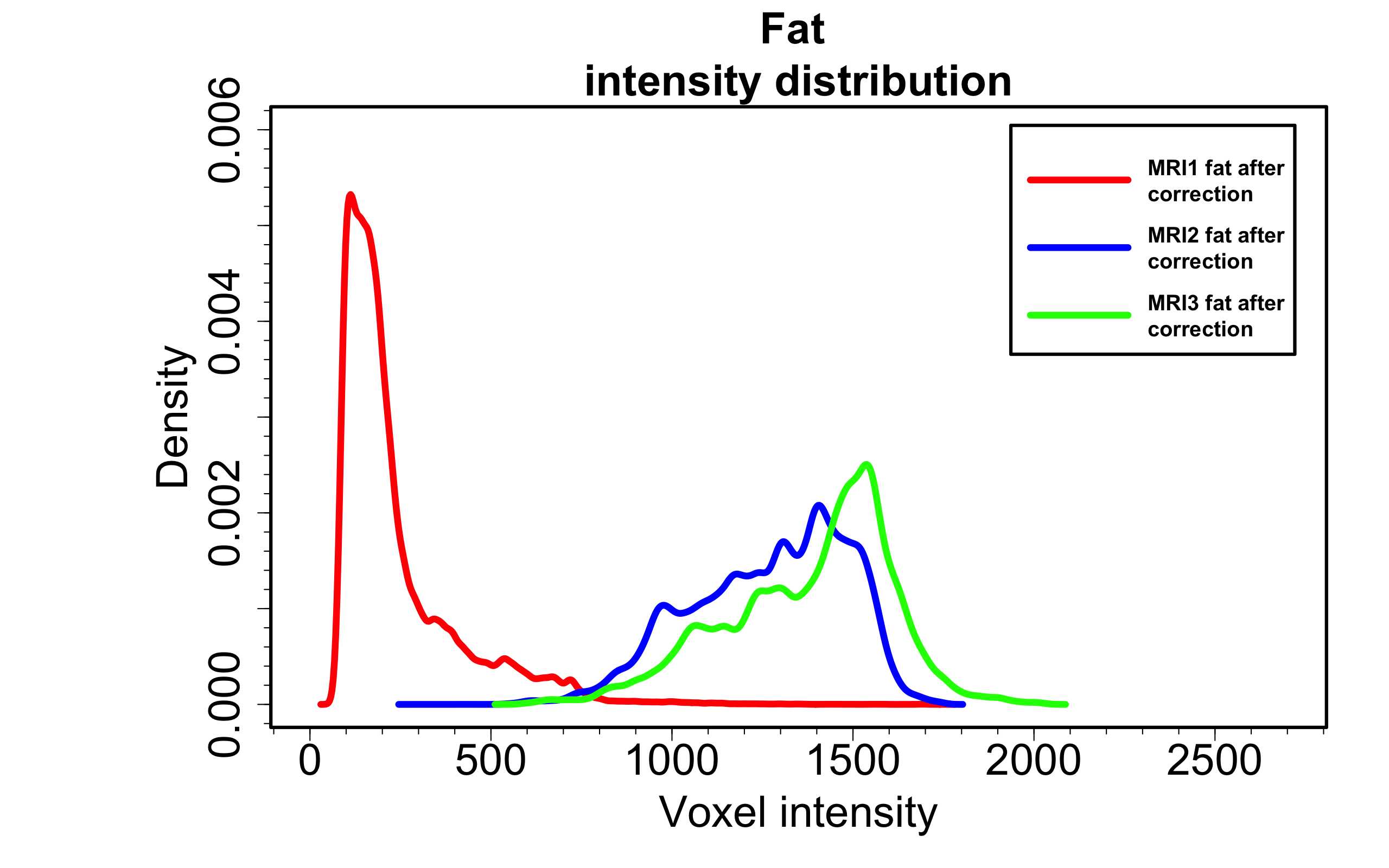}
\includegraphics[width=9 cm]{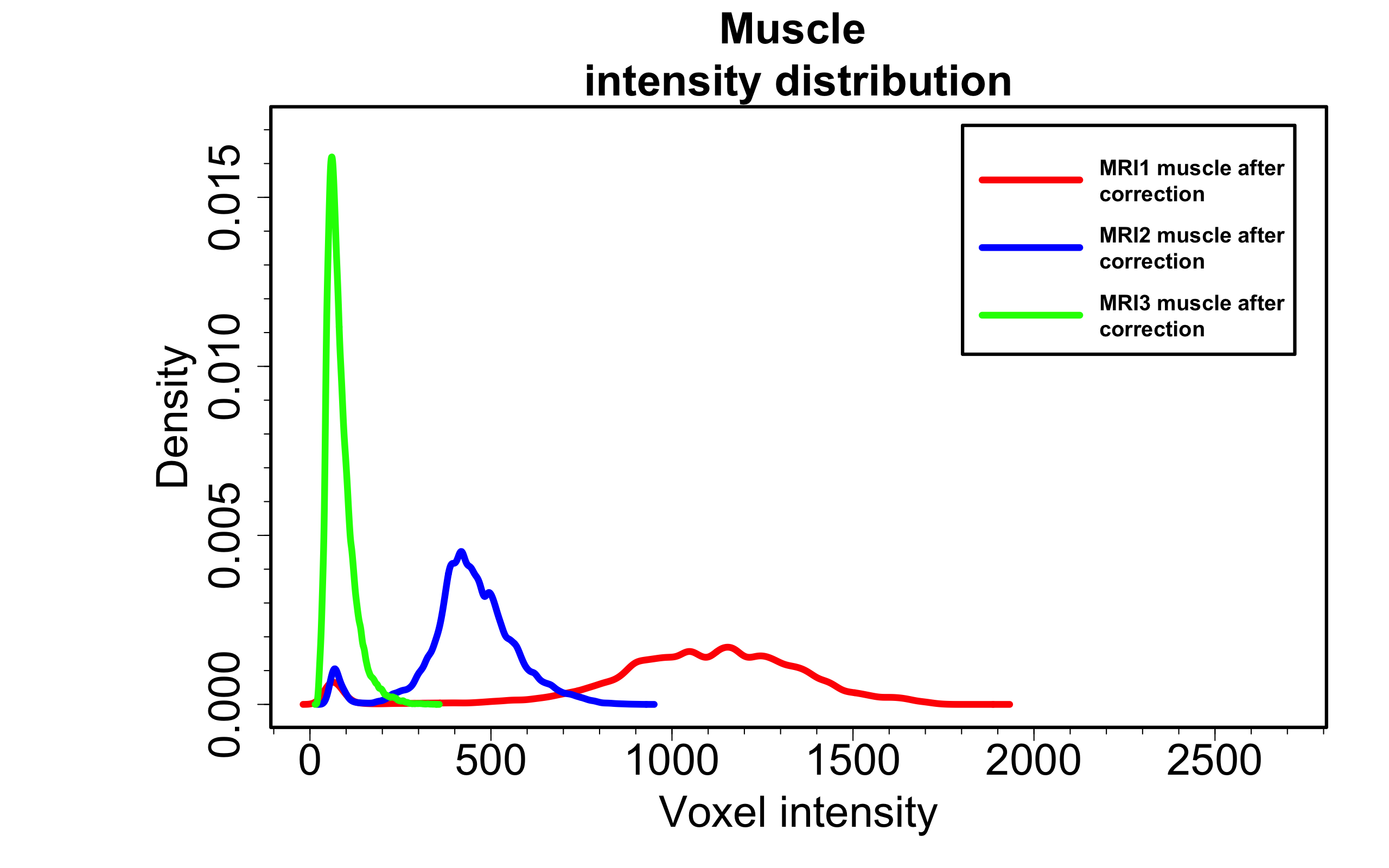}
\caption{Probability density estimations of fat and muscle intensities in water-only (MRI1), water-fat (MRI2), and fat-only (MRI3) images are shown. All images in the dataset (N=150) were used for this analysis. In the first row, although water-fat and fat-only have similar variances, the intensity distribution in water-only is quite different from the other two. In the second row, water-only, water-fat and fat-only have different distributions with different variances. Complementary information is noted therein, with water-fat contrast is better suited for Gaussian modeling, which is highly desirable in many statistical analysis settings. 
 \label{fig:hist}} 
\end{figure}

The FC algorithm is based on affinity functions. The most prominent affinities explained so far in the literature are adjacency-based ($\mu_d$), homogeneity-based ($\mu_\psi$), and object-feature based ($\mu_\phi$). In this work, we propose to combine different affinities of different MRI contrasts and utilize the combined affinity function set to conduct segmentation. Figure~\ref{fig:hist} shows fat and muscle intensity distribution (after images are corrected for noise and inhomogeneity) for water-only (MRI1), water-fat (MRI2), and fat-only (MRI3) images. As it is depicted, the standard deviations of intensities pertaining to muscle are quite different across different contrasts. Similarly, fat intensity patterns have different mean and variance across different image contrasts. It is also worth mentioning that fat-only and water-fat contrasts show large overlaps in fat histograms as would be expected for non-fat suppressed image types. However, infiltrated fat inside the muscle tissue can still be observed in the Figure~\ref{fig:hist} (first row, the difference between blue and green density distributions indicates infiltrated fat). For precise muscle analysis, it would be more appropriate to choose MRI2 as it follows a balanced Gaussian distribution which is easier to analyze and incorporate into segmentation and/or tissue characterization algorithms (Figure~\ref{fig:hist}, second row).

Based on all these qualitative observations and density histograms where intensity patterns follow distinct distributions, we propose to combine the affinity functions derived from each MRI contrast to quantify muscle and fat tissues jointly. Moreover, we also test the alternative hypothesis where individual segmentation from each image contrast will be combined as a decision fusion process. Here in this subsection, we confine ourselves to unifying affinity functions. In multi-contrast MRI of the thigh, let us have a fuzzy object membership affinity for each contrast: $\mu_\mathcal{O}^{MRI1}, \mu_\mathcal{O}^{MRI2},\mu_\mathcal{O}^{MRI3}$. Assume that there is a fuzzy object membership function $\mu_\mathcal{O}^{new}$ that combines the complementary strength of muscle and fat tissue intensity distributions from different MRI contrasts. One straightforward way to combine multiple affinities is the weighted summation approach as: $$\mu_\mathcal{O}^{new}=\sum_{i=1}^3 w_i \mu_\mathcal{O}^{MRI_i},\text{ s.t. } \sum_{i=1}^3 w_i=1,$$ where each affinity is weighted based on the effects of that particular MRI contrast on the final segmentation. Assuming that the segmentation accuracy is measured by dice similarity coefficient (DSC), then $w$ can be set as:
$$w_i=\frac{DSC(MRI_i)}{\sum_{i=1}^3DSC(MRI_i)}$$ \\
This step is conducted off-line, as once the weights are determined from each contrast based on the goodness of segmentation results; the system uses the same set of weights for the segmenting different images.

\subsection{Fusion of Multiple Segmentations (FC-Fusion)}
\begin{figure}
\centering
\includegraphics[width=6 cm]{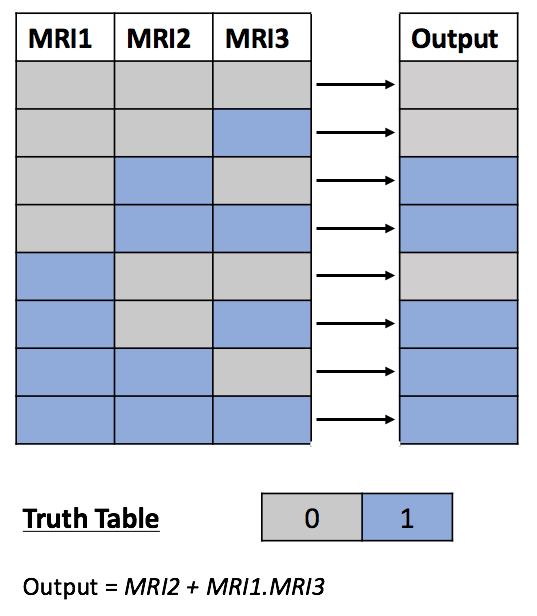}
\caption{Truth table for different combinations of multicontrast MRI. The final segmentation output is 1 either if MRI2 (water-fat) is available or MRI1 (water-only) and MRI3 (fat-only) are available. The simplified output is obtained using Karnaugh Map which gives FC-Fusion segmentation result. \label{fig:decisionfusion}} 
\end{figure}

As mentioned in the previous section, we also explore an alternative segmentation method based on fusion of multiple segmentations (namely, label or decision fusion). In decision fusion process, for designing FC fusion segmentation, we use a truth table for different input combinations. Figure~\ref{fig:decisionfusion} illustrates this process by indicating 0 or 1 for each entry of the truth label where 1 in output column indicates that both tissues (muscle and fat) can be recovered from the quantification process and 1 in left columns simply indicates if that particular image is used in fusion process or not. As can be seen from output column, to get fusion segmentation results, we need to use either MRI1 and MRI3 or sole MRI2 so as to maximize benefits of multi-contrast MRI. This indicates that, in the absence of MRI2, we need to use both MRI1 and MRI3 to generate the segmentation output. The truth table is simplified using a Karnaugh map (K-Map) to formulate the decision fusion and the resulting output map is determined as (MRI2 + MRI1.MRI3). The main reason for fusion of FC segmentation results is to compensate for any segmentation errors while performing segmentation over each modality separately.
\subsection{Automatic seed selection}
The proposed algorithm needs foreground and background seeds to initiate segmentation. Seed selection is critical in many medical image segmentation problems where literature is vast in proposing automatic seed selection procedures~\cite{pham2000current}. In this study, we devise three strategies for background and foreground seed sampling: manual, morphological filtering, and affinity propagation based seeding. We also evaluate these strategies' comparative performances in the proposed delineation framework.

\subsubsection{Manual seed selection}
Although our proposed method is fully automated, sometimes user interaction may be necessary for very challenging cases. In addition, it is often desirable to compare the fully automated method with interactive methods for efficiency purposes. Manual interaction is the most popular and perhaps the most reliable foreground and background seed sampling method. The strength of the manual seed selection comes from the superior recognition performance of human observers compared to automated methods. Sensitivity and specificity of manual seed selection in our experiments are 100\% as observers are not required to differentiate vague boundaries between fat and muscle. However, it is also important to note that manual seed selection is the localization/detection part of the segmentation process and it is completely different from manual annotation of the objects, which is often called ``manual segmentation". The drawback of this strategy is the computational overhead when multiple seeds are required to complete segmentation. As summarized in Table~\ref{table:efficiency}, manual seeding takes around 27 seconds, which is more than three times of the proposed segmentation's processing time. This duration may increase substantially for other methods, such as graph-cut segmentation where the user may need to put several seeds at locations where the segmentation algorithm is underachieving. 

\subsubsection{Morphological image filtering for seed generation}
Clustering and morphology filtering based methods have been shown successful in segmenting tissues. Due to the unique challenges of the problem at hand, it may be desirable to have more advanced methods than manual seeding. In our particular problem, clustering and morphology filtering based methods can identify tissue classes up to a certain level, even if they fail to segment whole tissue labeling. Figure~\ref{fig:overview} second row shows how this information can be obtained through morphology based filtering methods such as thresholding, erosion, and connected component analysis. Parameters of thresholding and connected component analysis are optimized for fat and muscle tissues only in water-fat images due to a better contrast between fat and muscle tissues. Once fat and muscle tissues are roughly identified by morphological filtering, we detect sample voxels for muscle and fat tissue locations from the largest connected components of the output. We also ensure that the sampled seeds have neighboring voxels which have the same label. This constraint avoids seed selection from uncertain locations (i.e., boundaries) which are vulnerable to mis-segmentations. 


\subsubsection{Affinity propagation for seed generation}
Affinity propagation (AP) is an unsupervised learning task which is used to partition data into meaningful and similar groups while considering the similarity between pairs of data points~\cite{frey2007clustering}. Although the AP clustering algorithm is used successfully in many areas like image segmentation~\cite{foster2014segmentation}, auto-detection of genes; automatic seed generation via AP clustering is a novel contribution. Moreover, the proposed seed generation methodology through AP is based on multi-objects (i.e., fat and muscle) on multiple MR image contrasts. Herein, we first estimate the probability density functions of multiple MR image contrasts and roughly localize fat and muscle regions by sampling the most discriminative intensity values (Figure~\ref{fig:overview}). To test the reliability of the sampled intensity locations, we use a connected component analysis to check the size of the object that the sampled voxel may belong to. Among the three MR image contrasts, experimental results and qualitative judgments show that water-fat and fat-only images provide the best contrast for seeding procedure which is described in the experiments and results section.

\section{Experiments and Results\label{sec:results}}
\subsection{Evaluation metrics}
We have evaluated the experimental results both qualitatively and quantitatively. For qualitative evaluations, two participating expert interpreters use their visual assessment to evaluate the segmentation results. Furthermore, we have used dice similarity coefficient (DSC) to evaluate how well the segmented regions overlap with the ground truth provided by the participating interpreters. Intra- and inter-operator agreement rates between expert interpreters were found to be 85.7\% and 83.5\%, respectively. We have also used the coefficient of variance (CoV) to determine how effectively intensity variations can be eliminated. Lower CoVs indicate improved quality in images due to pre-processing (i.e., inhomogeneity correction).

\subsection{Model evidence for MRI information content, and effects of pre-processing}

\begin{figure}[h]
\includegraphics[width=4.6 cm]{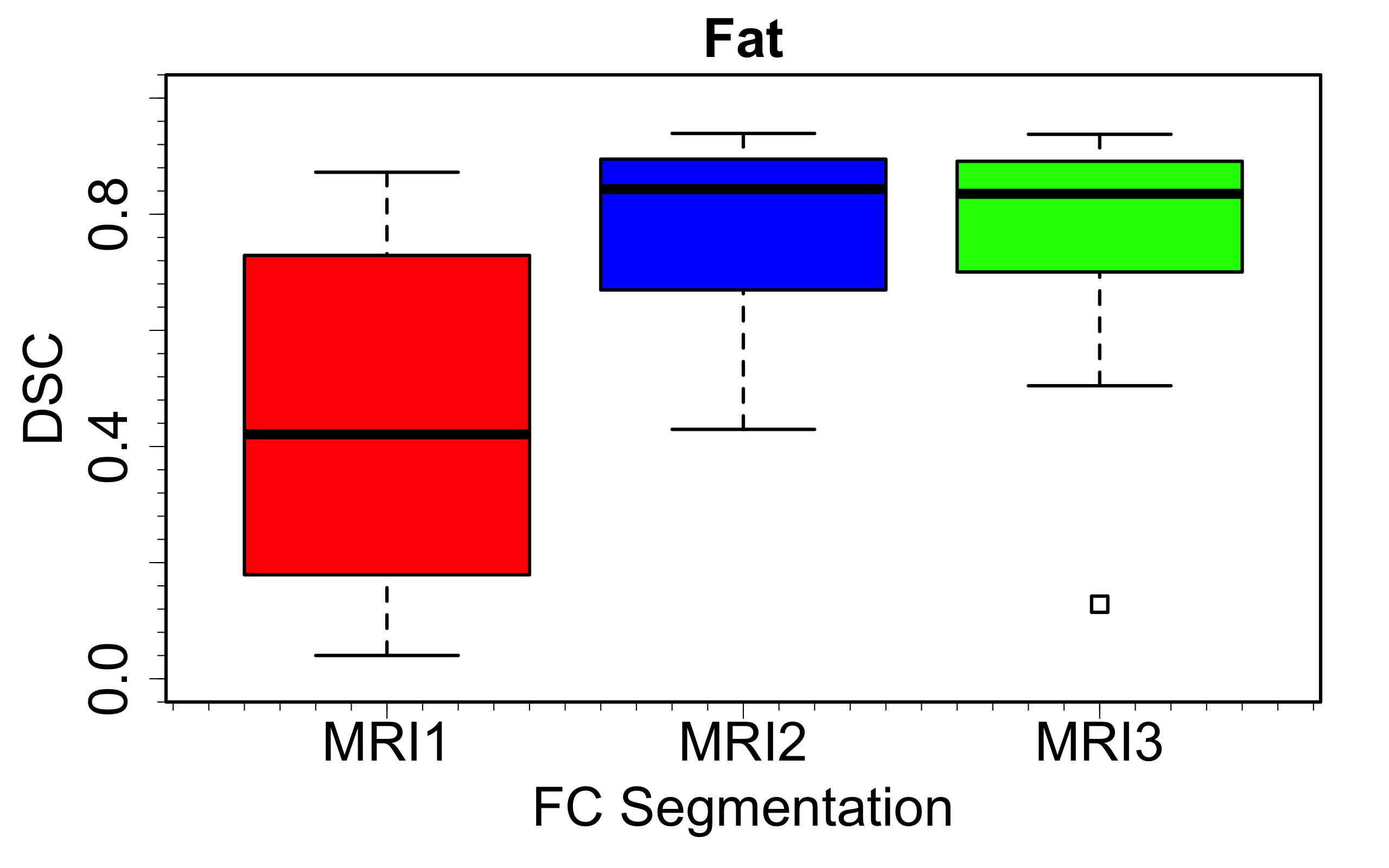}
\includegraphics[width=4.6 cm]{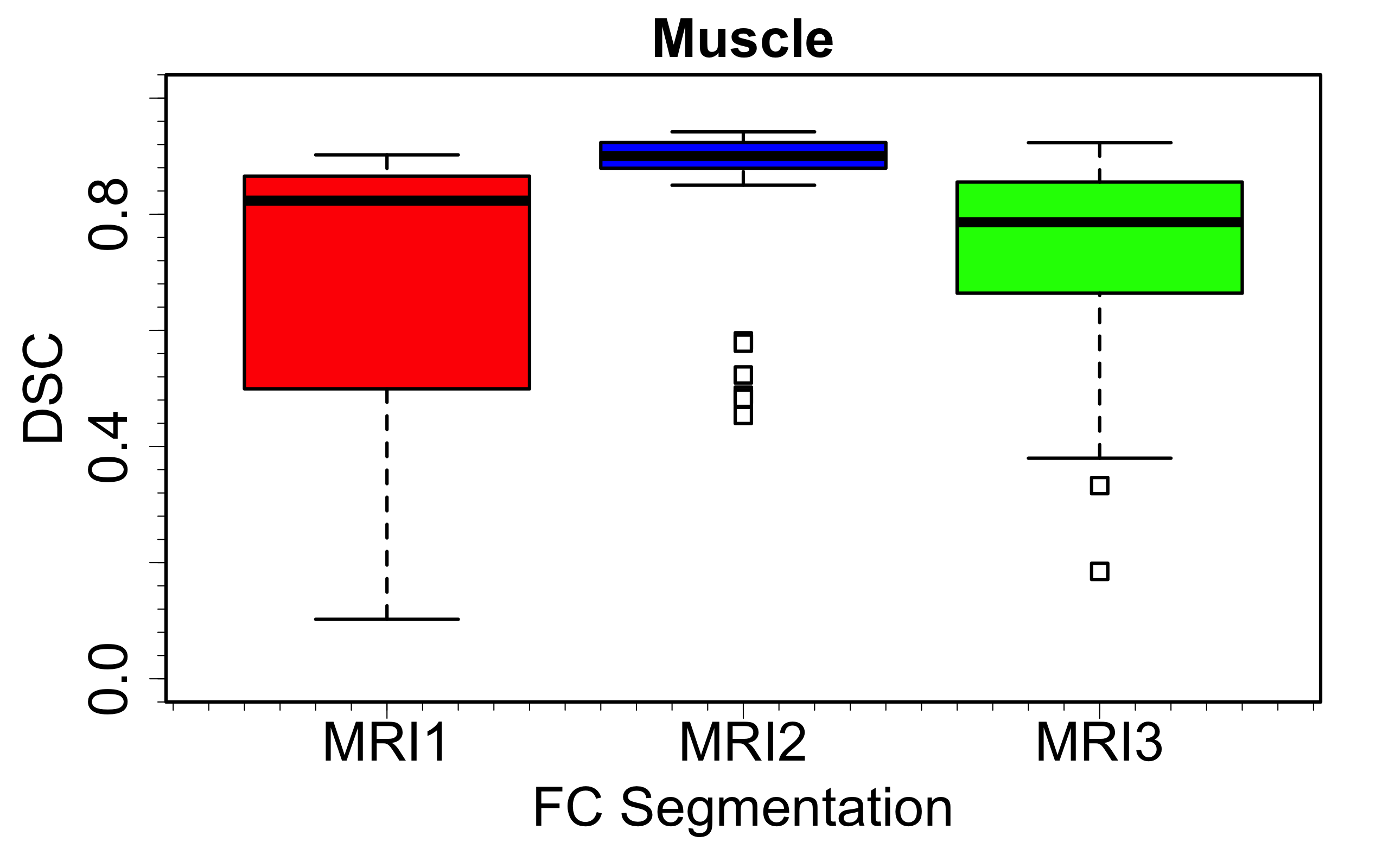}
\caption{Evidence obtained from the individual delineations of fat and muscle tissue show that highest accuracy was obtained when water-fat (MRI2) is used. Especially in muscle delineation, the standard deviation of the water-fat image is extremely small compared to the other modalities. Fat-only images (MRI3) gave similar accuracy as that of water-fat images in fat delineation but with larger variation. Water-only images (MRI1) seem to contain least discriminative information about the fat tissue. All images in the dataset (N=150) were used for this analysis.
\label{fig:evidences}} 
\end{figure}

To quantitatively study the information content of different MRI image contrasts, we calculate DSC pertaining to each image type. As shown in Figure~\ref{fig:evidences}, the highest DSC is obtained when water-fat (MRI2) is used. The lowest segmentation accuracy is obtained with water-only images (MRI1) as they contain the least discriminative information about the fat tissue. 

\begin{figure}[h]
\includegraphics[width = 8 cm]{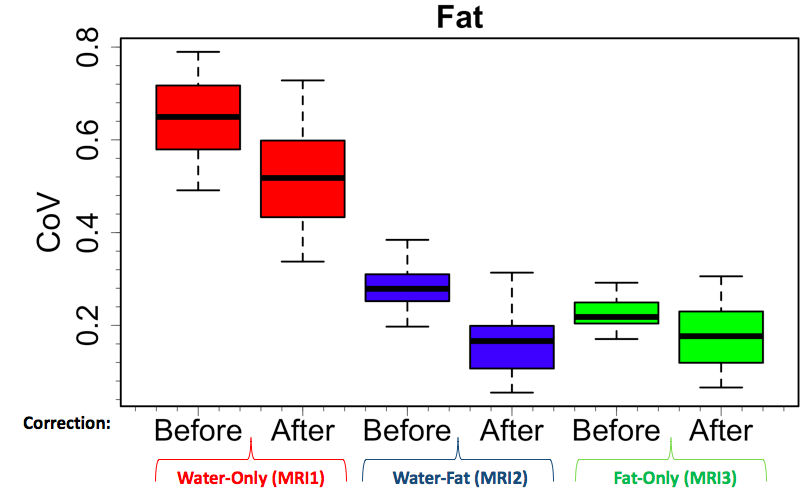}\\
\includegraphics[width = 8 cm]{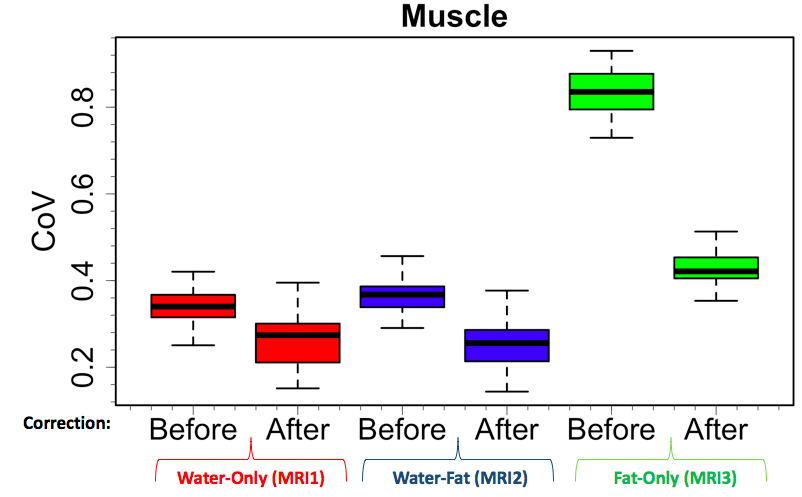}
\caption{Intensity variations prior to pre-processing steps and afterwards are represented as coefficient of variance (CoV). CoV of water-only (MRI1), water-fat (MRI2), fat-only (MRI3) images are summarized both for fat and muscle tissues, respectively. All images in the dataset (N=150) were used for this analysis.\label{fig:cov_comparision}} 
\end{figure}

It can be seen in Figure~\ref{fig:cov_comparision} that pre-processing makes CoV of fat and muscle tissues smaller. Since intensity variations are minimized prior to segmentation procedure with the pre-processing framework, segmentation quality is improved. In summary, the pre-processing will be substantially affecting the output not only for FC based segmentation methods but also for any segmentation approach that utilizes image intensities.

\subsection{Evaluation of the Segmentation Accuracy}
When segmentation accuracies are compared, the most important property is to have high mean accuracy with low standard deviation. We have compared our two proposed methods with the baseline FC segmentation as well (Table~\ref{table:dsc2}). Furthermore, we have quantitatively tested the effectiveness of the seeding procedure in terms of accuracy and robustness with respect to different seeding strategies. For automated seeding, we have proposed two procedures: AP and morphology based seeding strategies. The proposed method achieves the best results with high mean and low standard deviation as summarized in Table~\ref{table:dsc2}. All FC based segmentation results with the proposed AP seed generation are substantially better than other segmentation results, which are either based on manual or morphology based seed generation. In the proposed method, the segmentation with the automated AP seed generation gives the best DSC because the combination of weighted affinity functions further improved the segmentation quality. This is primarily due to the improved nonparametric modeling of the tissue class with AP clustering method.

\begin{figure*}[t]
\centering
\begin{subfigure}{.5\textwidth}
  \centering
  \includegraphics[width=0.95\linewidth]{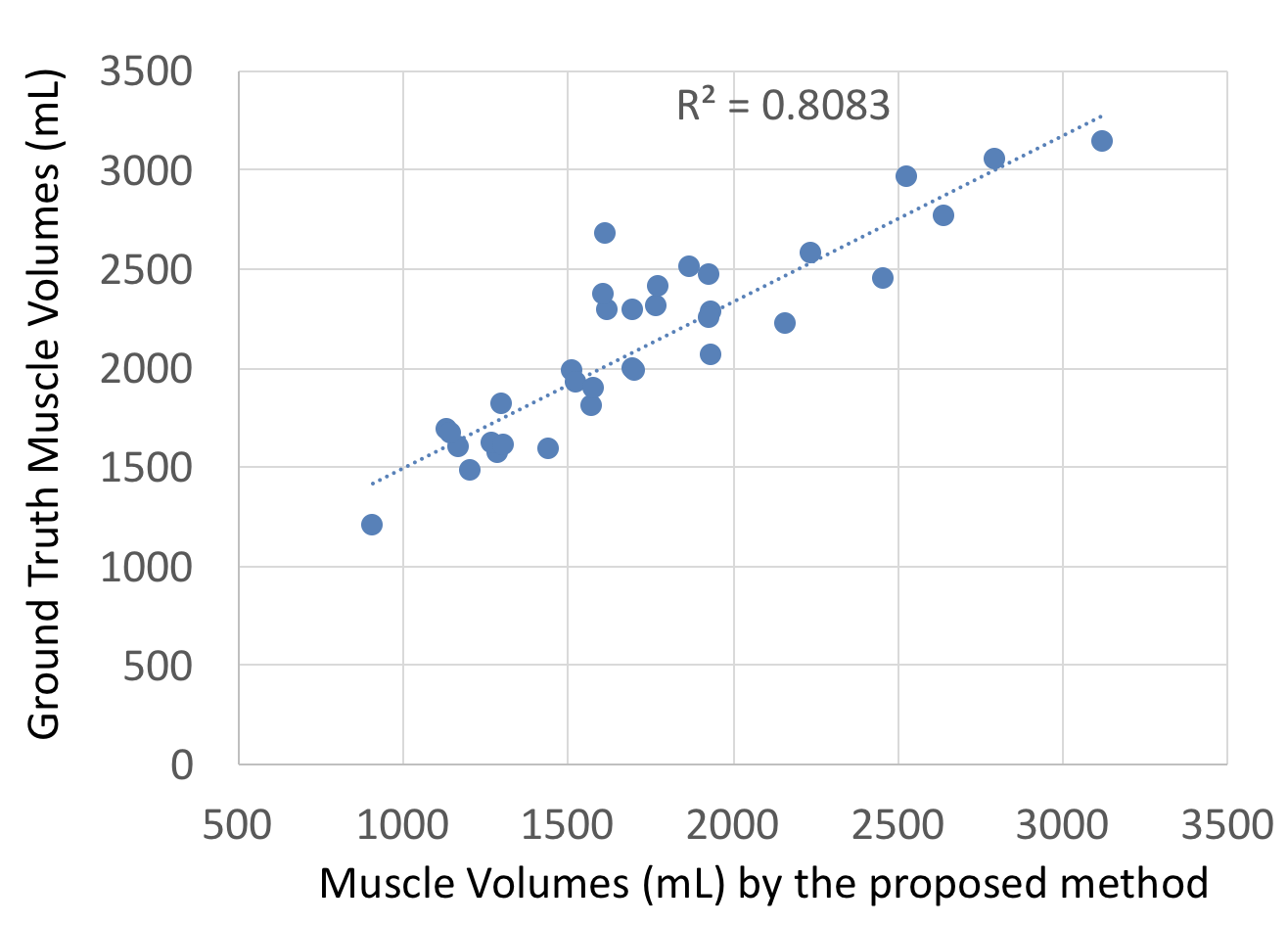}
  \label{fig:sub1}
\end{subfigure}%
\begin{subfigure}{.5\textwidth}
  \centering
  \includegraphics[width=1.0\linewidth]{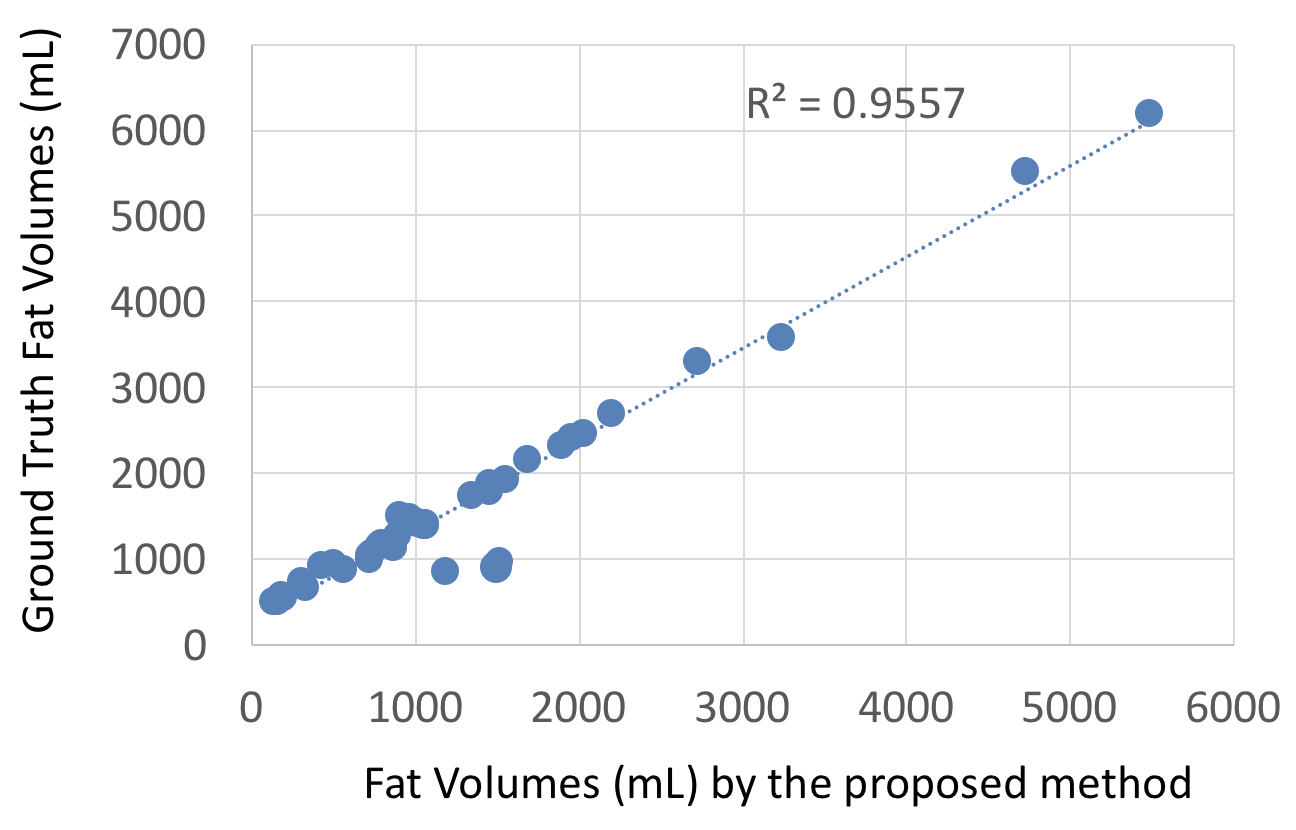}
\end{subfigure}
\caption{Correlation between muscle volumes and fat volumes with respect to ground truth measured volumes. }\label{fig:musfatvol}
\end{figure*}

Once segmentations are finalized, we have conducted a linear regression analysis for volumetric quantification of computed and true fat/muscle tissues. The correlation between the computed muscle-fat volumes and the ground truth volumes is shown in Figure~\ref{fig:musfatvol}. As it can be observed, the computed volumes are highly correlated with the ground truth volumes ($R^2 > 0.8$ and $R^2>0.95$ for muscle and fat volumes, respectively).
 
\begin{table}
\caption{Dice similarity coefficients (DSC) are listed for the proposed method(s). Two proposed methods (FC-Fusion and novel affinity based FC) are compared with the baseline method (FC segmentation) with respect to seeding methodologies. Higher DSC values indicate better segmentation accuracies. Novel affinity function based FC algorithm outperformed baseline and alternatively created algorithm (FC-fusion). The best seeding procedure was AP-seeding in all cases. All images in the dataset (N=150) were used for this analysis.} \label{table:dsc2}
\normalsize{
\begin{tabular}{lcc}
\\
\hline
\textbf{Method}              &      \textbf{Fat ($\%$)}               &    \textbf{Muscle ($\%$)}      \\ \hline
\textbf{FC Segmentation}              &            \textbf{DSC (mean/std)}        &         \textbf{DSC (mean/std)             } \\
Manual-Seeding                &  75.60 / 11.18  &  52.59 / 18.14\\
Morphology-Seeding        &  54.90 / 15.98  &   71.52 / 20.09\\
With AP-Seeding       &  77.26 / 11.57  &  72.88 / 18.59 \\ \hline
\textbf{FC-Fusion Segmentation}  &            \textbf{DSC (mean/std)}        &         \textbf{DSC (mean/std)}              \\
Manual-Seeding                &  77.51 / 17.46  &  82.07 / 9.27\\
Morphology-Seeding        &  77.39 / 18.84  &  84.99 / 17.64\\
With AP-Seeding       & 81.42 / 14.27   &  87.33 / 13.08\\ \hline
\textbf{Proposed Segmentation}    &            \textbf{DSC (mean/std)}        &         \textbf{DSC (mean/std)}              \\
Manual-Seeding                &  81.07 / 14.75  &  81.70 / 11.44 \\
Morphology-Seeding        &  80.66 / 10.36   &  85.27 / 20.36\\
With AP-Seeding       & 84.06 / 12.23   &  87.86 / 11.78 \\ \hline
\textbf{Proposed Segmentation}    &            \textbf{\% (mean/std)}        &         \textbf{\% (mean/std)}              \\
Sensitivity of AP-Seeding & 79.27 / 12.46 & 84.18 / 16.75 \\
Specificity of AP-Seeding & 98.16 / 8.50 & 96.78 / 10.10\\ \hline 
\end{tabular}}
\end{table}

\subsection{Comparison to other methods}
As briefly summarized in Section I-B, available approaches for segmentation of fat and muscle for thigh MRI can be categorized under boundary based methods~\cite{makrogiannis2016image}, region based methods such as Markov random field (MRF)~\cite{mrf}, graph-cut~\cite{chen2012medical} and clustering~\cite{tan2016detection}, and machine learning based methods such as random (decision) forest (RF)~\cite{randomforest}, multilayer perceptron (MLP)~\cite{backprop}, SVM~\cite{svm}, BayesNET~\cite{bayesnet}, Hoeffding trees~\cite{hulten2001mining}, Random committee~\cite{lira2007combining}, and AdaBoost~\cite{adaboost}. We compare our proposed method with several existing methods both qualitatively and quantitatively. Table~\ref{table:dsc} lists the quantitative comparisons of the proposed method's DSCs for fat and muscle segmentation with Boykov's graph-cut~\cite{chen2012medical}, active contour~\cite{makrogiannis2012automated,makrogiannis2016image} and machine learning based methods of random forest~\cite{randomforest} and random committee~\cite{lira2007combining}.

We report statistical validation of FC based (proposed) and other methods' segmentation using DSC in Table~\ref{table:dsc}. Proposed FC based algorithm showed higher DSCs compared to other methods. Among the widely used methods, Boykov's Max Flow Min Cut works better than active contour for fat segmentation but worse for muscle segmentation. While analyzing the classifier based segmentation results, although random forest and random committee show quite similar DSCs, random committee is a little better as it has higher mean and lower deviation. Among the proposed FC based-methods, the best results were obtained when fusion was conducted at the affinity function level: arranging the weight of the affinity functions for each imaging modality (i.e., image contrast of the MRI) significantly improved the DSC for fat and muscle to around 84\% and 87\%, respectively (Table~\ref{table:dsc2}). The results from our proposed and alternative segmentation methods prove the efficacy and feasibility of our proposed solution.

\begin{figure*}
\includegraphics[width=18 cm]{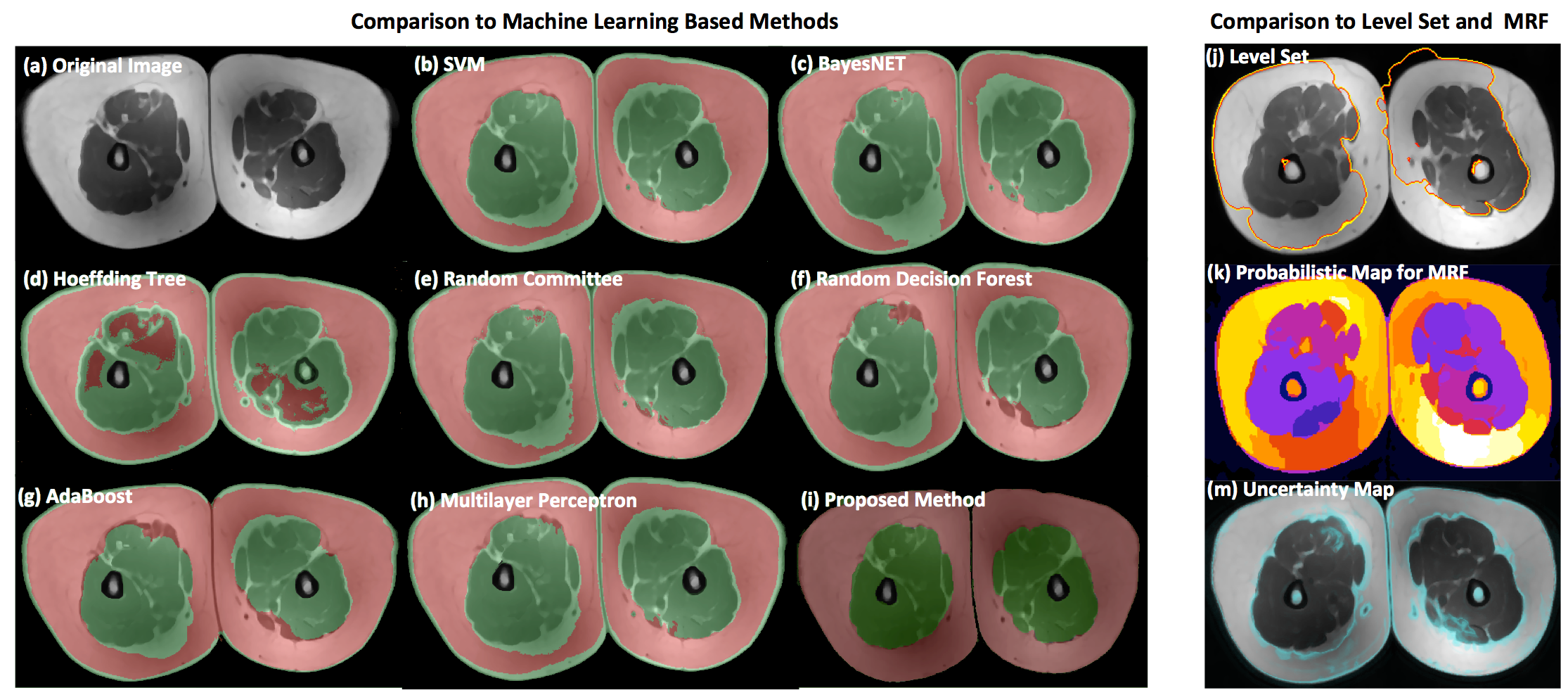}
\caption{Comparison of fat and muscle tissue segmentation results for a given MRI slice (a). Machine learning algorithms: SVM in (b)~\cite{svm}, BayesNET in (c)~\cite{bayesnet}, Hoeffding tree in (d)~\cite{hulten2001mining}, Random committee in (e)~\cite{lira2007combining}, Random forest in (f)~\cite{randomforest}, AdaBoost in (g)~\cite{adaboost} and Multilayer perceptron in (h)~\cite{backprop} are trained separately and optimized results are used for comparison with respect to the proposed algorithm (i). The selected MRI slice was used to qualitatively illustrate the impact of image inhomogeneity and fat tissue distribution on segmentation accuracy of each algorithm; the proposed approach (i) shows a dramatic improvement (no over- or under-segmentation issue observed) compared to all other methods. Other segmentation methods such as level set (j)~\cite{lim2013introducing} and MRF (k)~\cite{mrf} are also illustrated. The uncertainty map of the given slice (calculated based on homogeneity of the tissue distribution during FC formulation) indicates potential locations where an algorithm may fail (m). 
\label{fig:qualitativecomparison}}
\end{figure*}

For qualitative comparison with other methods, Figure~\ref{fig:qualitativecomparison} illustrates an example of the segmentation performance for several different methods. We have selected a representative image slice from a single image contrast of the thigh MRI in Figure~\ref{fig:qualitativecomparison}(a), and segmentation results of region based, boundary based, and machine learning based methods are shown in Figure~\ref{fig:qualitativecomparison}(b-m). The proposed algorithm shows improved segmentation results in Figure~\ref{fig:qualitativecomparison}(i) compared to others. Several of the other methods were not able to capture large fat regions in the subcutaneous compartment while leakages (i.e., over-segmentation to other tissues) were inevitable in most cases.

\subsection{Feasibility Studies: Brain and Whole-Body Tissue Segmentation}
\begin{figure*}[t]
  \centering
  \includegraphics[width=15 cm]{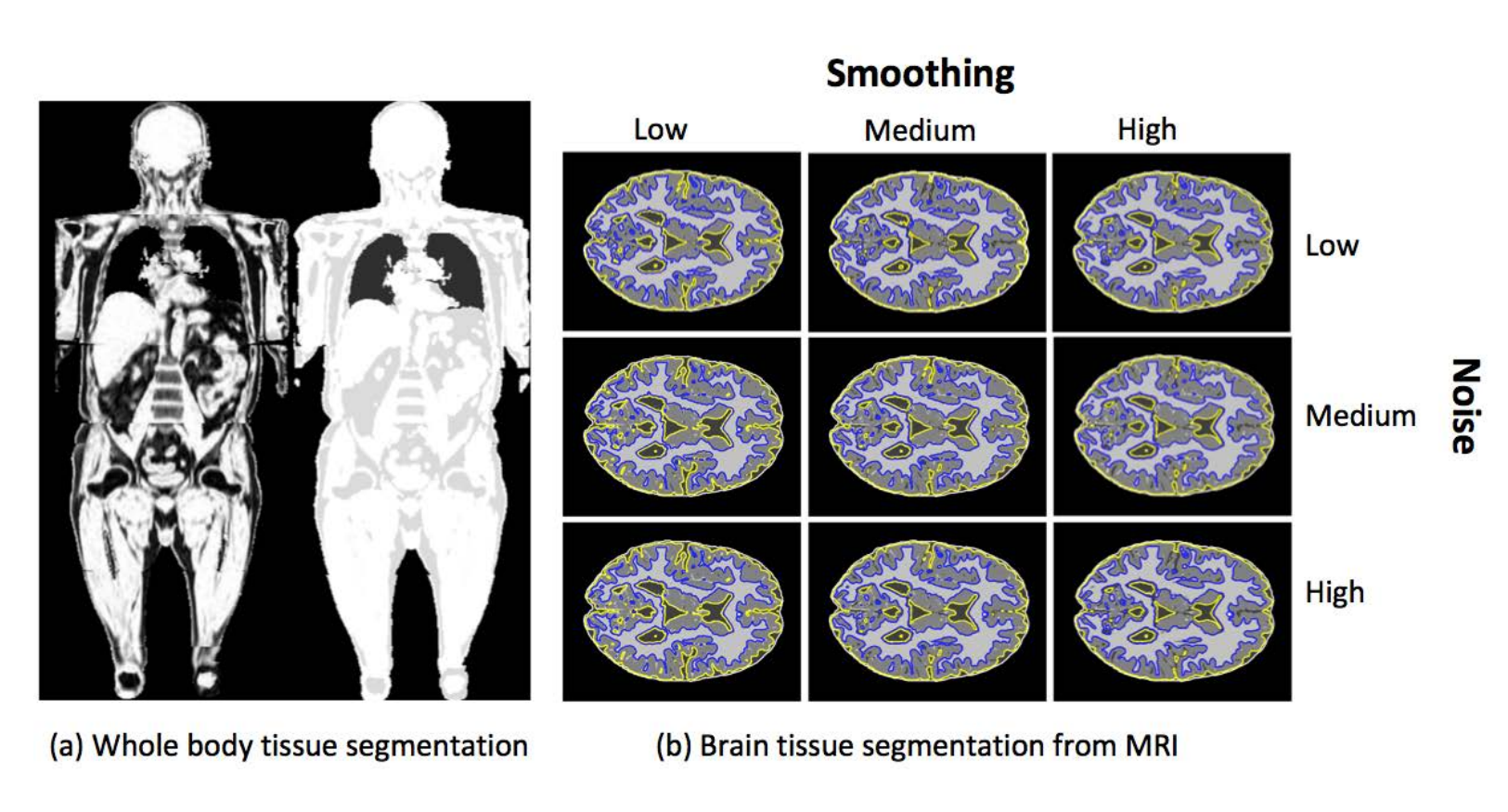}
  \caption{The generalizability of the proposed approach is shown via (a) whole-body tissue segmentation and (b) brain tissue segmentation. The whole-body tissue segmentation is performed on Dixon MRI sequences, whereas in (b), brain tissue segmentation with varying levels of noise and smoothing is depicted.}
  \label{fig:segment}
\end{figure*}

To evaluate the generalizability of our proposed approach both for different MRI type and body regions, we performed two additional experiments. In the first experiment, we applied our algorithm into Dixon MRI sequences for whole body tissue segmentation. Briefly, for a given sample whole-body MRI (Dixon, Figure~\ref{fig:segment}(a)-left), its 3-class tissue segmentation (Figure~\ref{fig:segment}(a)-right) was obtained within seconds. In our implementation, we considered fat, muscle, and lung fields in the segmentation. Resulting tissue delineations were visually evaluated (and found to be clinically feasible) by the participating experts of this study.

In the second experiment, our proposed method was validated using 20 simulated brain MRI scans (T1-weighted)~\cite{mriphantom1}, with the ground truth known. Various noise and smoothing levels were also used to evaluate the robustness of the proposed method. Three smoothing levels (small, medium, and large) were applied to the images using a Gaussian smoothing filter. Three noise levels (small, medium, and large) were applied to the images as well. With the proposed approach, considering all 20 brain MRI scans, Gray Matter (GM) and White Matter (WM) segmentation accuracies were excellent with the sensitivity and specificity rates above 90\%; Cerebrospinal fluid (CSF) segmentation was good with the sensitivity and specificity always above 85\%. The literature demonstrates that CSF is challenging to segment, with sensitivities generally in the 80\% - 88\% range and specificities between 51\% to 88\%~\cite{mriphantom2}. Figure~\ref{fig:segment}(b) shows segmented brain tissues. These feasibility results (both whole body and brain tissue segmentations) demonstrate the promising generic aspects of our proposed method in segmentation and quantification of different body regions, and type of MRI scans. The segmentation results are shown in Table~\ref{table:other_organs}.\\

\subsection{Computational efficiency}
Proposed segmentation and seeding (AP) methods have achieved the lowest computational time while guaranteeing higher accuracy as listed in Table~\ref{table:efficiency}. The slowest segmentation procedure took only 57 seconds, of which 27 seconds were spent on seeding and 30 seconds were used for fusion strategy. The shortest segmentation duration was 10.33 seconds provided by the proposed AP-seeding based FC segmentation algorithm integrating adjusted affinities. Manual seeding was restricted to a real clinical setting where a few seeds were often desired to initiate the segmentation; otherwise, seeding process itself can be time-consuming. Increasing the number of seeds led to similar segmentation results as that obtained through AP based seeding method but at the expense of an increase in the analysis time.

\begin{table}
\caption{Dice similarity coefficients (DSC) are listed for the proposed and alternative methods to show their efficacy in delineating fat and muscle tissues. Higher DSC values indicate better segmentation accuracies. All algorithms were reproduced, optimized, and applied to our data for a fair comparison. All images in the dataset (N=150) were used for this analysis.} \label{table:dsc}
\normalsize{
\begin{tabular}{lcc}
\\
\hline
\textbf{Method}              &      \textbf{Fat ($\%$)}               &    \textbf{Muscle ($\%$)}      \\ \hline
\textbf{Graph-Cut}              &            \textbf{DSC (mean/std)}        &         \textbf{DSC (mean/std)}              \\
Boykov's MaxFlow~\cite{chen2012medical}                &  73.50 / 16.75  &  72.20 / 15.01 \\ \hline
\textbf{Active Contour~\cite{makrogiannis2012automated,makrogiannis2016image}}              &            \textbf{DSC (mean/std)}        &         \textbf{DSC (mean/std)}              \\
Active contour/snake              &  71.41 / 16.13  & 74.73 / 14.86 \\ \hline
\textbf{Clustering}              &            \textbf{DSC (mean/std)}        &         \textbf{DSC (mean/std)}              \\
Random Forest~\cite{randomforest}                & 73.41 / 15.38   & 71.19 / 18.41  \\
Random Committee~\cite{lira2007combining}                &  74.83 / 13.39  & 71.90 / 16.42 \\ \hline
\textbf{Proposed Segmentation}    &            \textbf{DSC (mean/std)}        &         \textbf{DSC (mean/std)}              \\
With AP-Seeding       & 84.06 / 12.23   &  87.86 / 11.78 \\ \hline
\end{tabular}}
\end{table}

\begin{table*}
\begin{center}
\caption{Segmentation performance of the proposed approach for whole body and brain tissue segmentation.} \label{table:other_organs}
\normalsize{
{\color{black}\begin{tabular}{lcccc}
\\
\hline
\textbf{Tissues}              &      \textbf{Number of Images}     &      \textbf{Dice Score}          &    \textbf{Sensitivity}  &    \textbf{Specificity }    \\ \hline
\textbf{Brain Tissue}          &      20             &       \textbf{-}  & 	\textbf{-} 	& \textbf{-}               \\
Gray Matter         &       \textbf{-}            &  0.93          & 	0.92	&	 0.94          \\
White Matter         &        \textbf{-}           &       0.93 & 0.95 & 0.91               \\
\hline
\textbf{Whole-Body}            &   3                &        \textbf{-}  & 	\textbf{-} 	& \textbf{-}           \\
Fat         &        \textbf{-}           &   0.90       & 0.89 &    0.99        \\
Muscle         &         \textbf{-}          &   0.88      & 0.87      & 0.99       \\
\hline
\end{tabular}}}
\end{center}
\end{table*}

\begin{table}
\caption{Computational efficiencies of the proposed framework's subsequent steps (seeding and segmentation) are summarized for fat and muscle tissues, respectively.} \label{table:efficiency}
\normalsize{
\begin{tabular}{lrrr}
\\
\hline\hline 
\textbf{Method}              &      \textbf{Fat}               &    \textbf{Muscle}      & \textbf{Total} \\ \hline
FC Segmentation              &             3   sec        &         5 sec            &  8 sec   \\
FC-Fusion Segmentation  &            15  sec         &        15 sec               & 30 sec\\
Proposed Segmentation    &             3   sec         &         5 sec               & 8 sec\\\hline\hline
Manual-Seeding                &  	          12 sec                &        15 sec                & 27 sec\\
Morphology-Seeding        &  	          1 sec                 &           2  sec           & 3 sec\\
Proposed AP-Seeding       & 	       1.17    sec                 &        1.16 sec                & 2.33 sec\\\hline
\end{tabular}}
\end{table}

\subsection{Quantification from a single image slice}
\begin{figure*}[t]
\centering
\begin{subfigure}{.5\textwidth}
  \centering
  \includegraphics[width=0.97\linewidth]{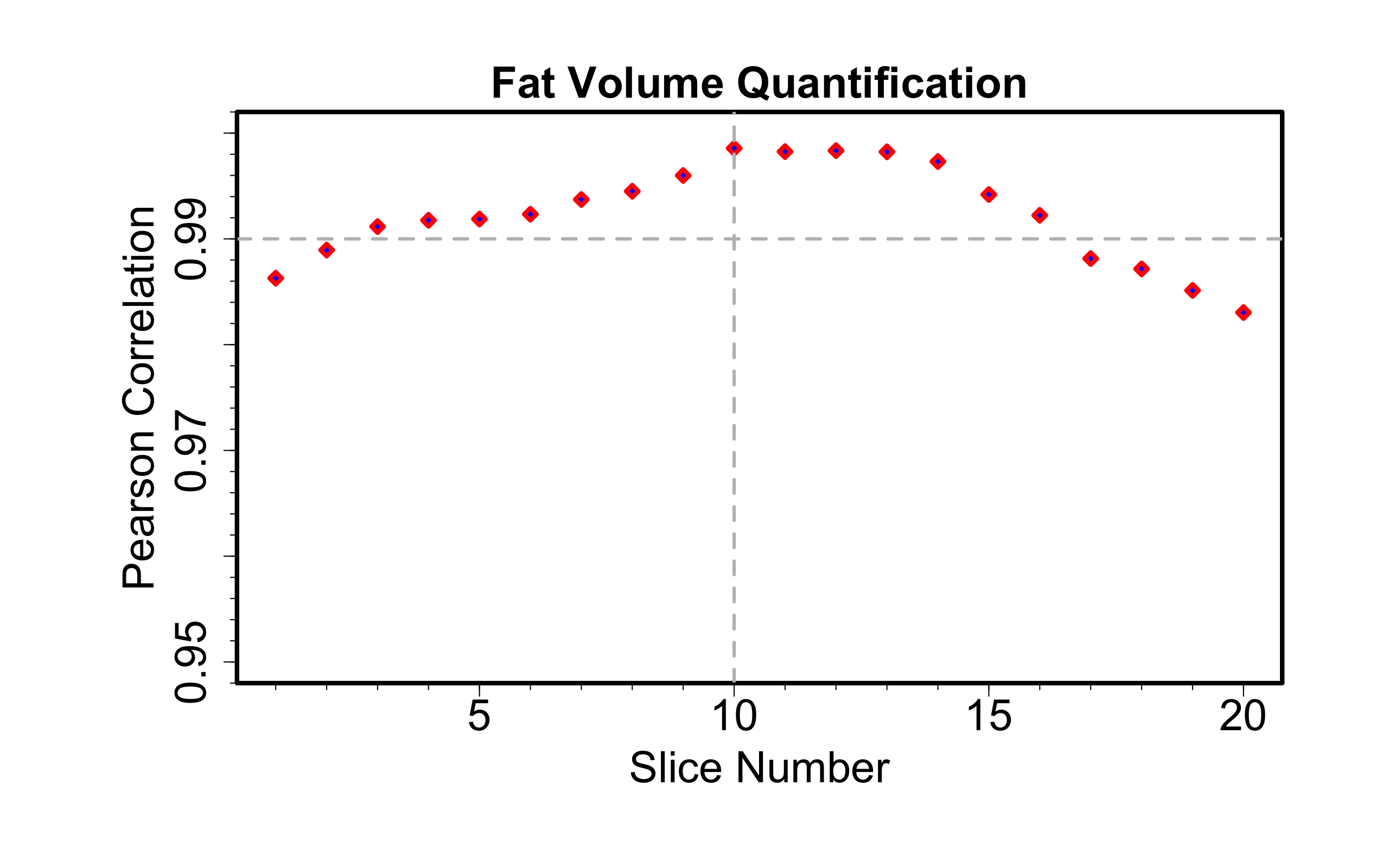}
  \label{fig:sub1}
\end{subfigure}%
\begin{subfigure}{.5\textwidth}
  \centering
  \includegraphics[width=1.0\linewidth]{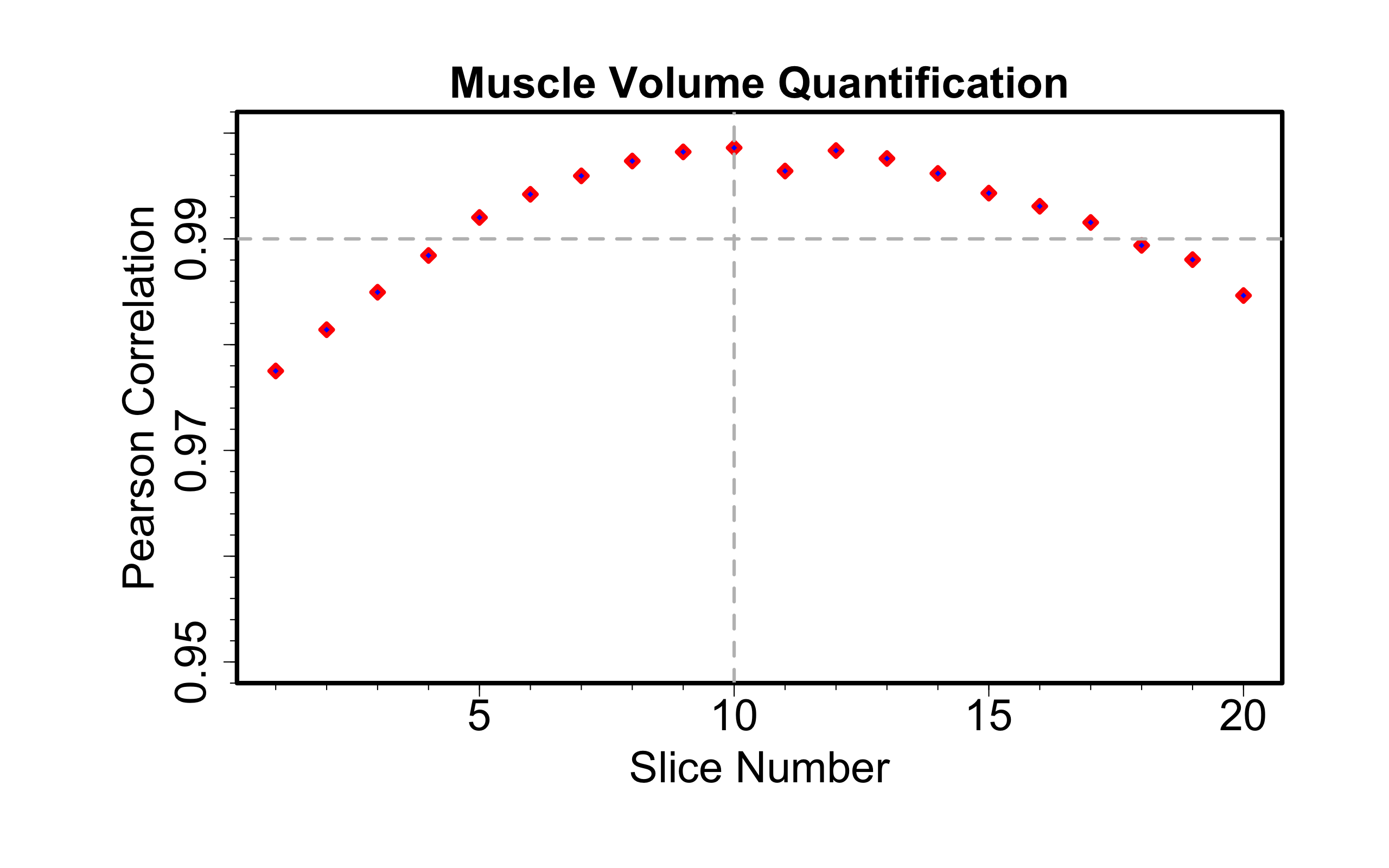}
\end{subfigure}
\caption{Fat and muscle volumes are correlated with a single slice from its stack. Pearson correlation is maximized when the single slice is in the mid-thigh region.\label{fig:singleslice}}
\end{figure*}

In clinical practice, radiologists frequently use a single slice to estimate the whole volume for a region in an object. For instance, in abdominal fat quantification, instead of segmenting the whole abdomen and separating subcutaneous tissue from visceral, which is a challenging problem, a single slice at the umbilical level is often selected and used to quantify abdominal fat volume~\cite{sarfaraz,nmc}. The main reasons for this practice are two-fold. First is a lack of software algorithms that can work efficiently and accurately. This is because many algorithms fail to provide exact quantification, and thus user interaction is necessary for a better quantification. It is time-consuming for radiologists to correct each segmentation failure for volumetric images. Second, the anatomy does not change abruptly from slice to slice. Thus, it is a reasonable assumption that the content of the tissue distribution in a single slice will be correlated with the whole volume. Then, based on a known correlation between a single slice and volume, creating a linear regression equation will estimate the volumetric tissue distribution from a single slice. Inspired by this clinical practice and to explore the inter-relationship of thigh MRI slices, we perform a Pearson correlation test between fat and muscle volume extracted from a single slice with its volumetric surrogate truths. Figure~\ref{fig:singleslice} shows correlation value for each slice when estimating the volumetric distribution of fat and muscle, respectively. As indicated, each slice contributed more than $R=0.97$ correlation value, and the highest correlations $R>0.99$ were obtained when mid slice of the thigh region was selected. This indicates that the proposed system can be used for much faster muscle and fat volume estimation from a single slice for practical purposes.

\section{Discussions and Concluding Remarks}
\label{sec:discussion}
\subsection{Limitations and Future Works}
Our study has a few limitations to be noted. With its success in almost all tasks of visual domain and its extensive use for medical image analysis, deep learning can serve as an alternative to our proposed approach. The joint segmentation of multi-contrast images in deep neural network framework can lead to accurate and robust performance. In addition, the parameters in our proposed approach were learned offline, which can be estimated in an online manner for reliable performance. Moreover, in the context of skeletal muscle composition studies, more advanced MRI acquisition methods like Dixon imaging can be used which provide an improved correspondence between the image signal and the fat and water fractions within a given voxel. Lastly, it is worth mentioning that partial volume can be an issue to be solved for separating water and fat signals when their contributions are roughly equal at a given voxel. Especially when the resolution is limited, this effect should be corrected. Although we did not observe large errors due to partial volumes, conventional partial volume correction methods for MR images such as~\cite{pvc} can be applied to minimize such errors in volumetric quantification. Furthermore, Dixon can be used for fat/muscle quantification in sub-voxel level unlike our method where we used voxel level quantification.

Our future study will involve exploring another important imaging marker, called biomechanical muscle quality (BMQ), for a potential association of aging and age-related metabolic diseases. In order to calculate BMQ, we will identify the intra-muscular adipose tissue (IMAT), which is associated with a reduction in force production and suggested to affect BMQ. For this purpose, a unified framework to perform segmentation and quantification of the three tissues (muscle, fat, and IMAT) can be utilized in the clinical evaluation of BMQ. The current work also suggests a potential for application of deep learning based methods for improved multi-contrast image segmentation approaches. Inspired by the recent success of purely machine learning methods and the current methods' limited achievement on IMAT quantification (i.e., around 70-80\% success bands~\cite{makrogiannis2012automated,makrogiannis2016image,tan2016detection}), we aim to explore the joint segmentation of these three tissues using a deep learning framework in our future study. Moreover, given a limited number of images, we performed evaluations on a small dataset for whole-body tissue classification as well as brain tissue segmentation. Although the results are quite promising, with a larger dataset, deep learning based approaches can be explored for improved segmentation of whole-body and brain tissues.

\subsection{Summary}
Conventional methods for the tissue delineation and quantification mostly rely on manual or semi-automated algorithms, having low accuracy and efficiency. In the case of multi-contrast MRI scans, conventional approaches use isolated frameworks for each imaging modality; thus, leading to suboptimal results. Given the clinical importance of fat and muscle segmentation and quantification in thigh MRI, as well their utility in other research settings, we proposed a novel approach to address unique challenges associated with these images and showed the generalization ability of the proposed system for the whole-body tissue composition and the brain tissue delineation. We provided a complete quantification framework that automatically minimizes extreme challenges (inhomogeneity, noise, and non-standardness) of the MRI automatically. To maintain the unsupervisory and automated nature of our proposed approach, we performed automatic seed generation using an affinity propagation clustering algorithm. In the delineation step, to the best of our knowledge, the proposed approach is the first to explore the utility of multiple affinity functions to perform a unified segmentation of multiple objects from multi-contrast MRI. Our automated seed selection approach also evades the need for filtering approaches which can be prone to subjectivities.


\section*{Acknowledgment}
This work was supported in part by the Intramural Research Program of the National Institute on Aging of the National Institutes of Health (NIH).


\bibliographystyle{IEEEtran}
\bibliography{thigh} 
\end{document}